\documentclass[11pt]{article}

\usepackage{acl}

\usepackage{times}
\usepackage{latexsym}

\usepackage[T1]{fontenc}

\usepackage[utf8]{inputenc}

\usepackage{microtype}

\usepackage{inconsolata}

\usepackage{graphicx}

\usepackage{amsmath}
\usepackage{amssymb}
\usepackage{mathtools}
\usepackage{amsthm}

\title{Large Vocabulary Size Improves Large Language Models}

\author{Sho Takase \hspace{1.5em} Ryokan Ri \hspace{1.5em} Shun Kiyono \hspace{1.5em} Takuya Kato \\
  SB Intuitions \\
  {\tt \{sho.takase, ryokan.ri, shun.kiyono, takuya.kato\}@sbintuitions.co.jp} \\
  }

\begin{document}
\maketitle
\begin{abstract}
This paper empirically investigates the relationship between subword vocabulary size and the performance of large language models (LLMs) to provide insights on how to define the vocabulary size.
Experimental results show that larger vocabulary sizes lead to better performance in LLMs.
Moreover, we consider a continual training scenario where a pre-trained language model is trained on a different target language.
We introduce a simple method to use a new vocabulary instead of the pre-defined one.
We show that using the new vocabulary outperforms the model with the vocabulary used in pre-training.
\end{abstract}

\section{Introduction}


Since the GPT series demonstrated that Large Language Models (LLMs) excel in complex reasoning tasks~\cite{radford2018improving,noauthororeditor,NEURIPS2020_1457c0d6}, they have rapidly become indispensable tools for various natural language processing tasks.
To construct better LLMs, previous studies have addressed theoretical analyses of internal layers~\cite{DBLP:conf/icml/XiongYHZZXZLWL20,takase2024spike} and conducted extensive experiments to provide empirical findings~\cite{kaplan2020scaling,Hoffmann2022TrainingCL,wortsman2024smallscale}.
For example, \citet{Hoffmann2022TrainingCL} reported the compute-optimal training configuration, which determines suitable parameter and training data sizes for a given computational resource.

In contrast, although previous studies have explored the properties of internal layers in LLMs, parameters related to the vocabulary, the embedding and output layers, are under-explored.
Specifically, there are no well-established findings on how to determine the subword vocabulary size, which defines the parameter size of the embedding and output layers.
As a standard strategy, a vocabulary size in the 30k-60k range is used for monolingual LLMs~\cite{noauthororeditor,NEURIPS2020_1457c0d6,black2022gptneox20b,zhang2022opt,touvron2023llama}, while around 250k is used for multilingual LLMs~\cite{chowdhery2022palm,le-scao-etal-2022-language}.
For monolingual LLMs, a larger vocabulary size has been discussed in terms of efficiency during the inference phase~\cite{almazrouei2023falcon}.
However, the question remains: does a larger vocabulary size offer any advantages for the quality of monolingual LLMs?
To address this question, we empirically investigate the relationship between vocabulary size and performance on downstream tasks.

We conduct experiments on two languages: English, which is widely used, and Japanese, which is character-rich.
We show that a larger vocabulary size improves the performance of LLMs in both languages.
In addition to training from scratch, we consider the continual training scenario.
When adapting a pre-trained LLM to another language, it may be beneficial to reconstruct an appropriate vocabulary instead of reusing the original vocabulary.
For this purpose, we propose a strategy to swap parameters related to the vocabulary.
We demonstrate that using the reconstructed vocabulary can improve performance.

\section{Vocabulary Construction}
\label{sec:vocab_construction}
To construct subword vocabularies, there are two widely used algorithms: Byte-Pair Encoding (BPE)~\cite{sennrich-etal-2016-neural} and unigram language model~\cite{kudo-2018-subword}.
In this study, we use the unigram language model implemented in SentencePiece~\cite{kudo-richardson-2018-sentencepiece}.
For each language, we use the following vocabulary sizes: 5k, 10k, 50k, 100k and 500k.


We conduct experiments on two languages: English and Japanese.
For the English training data, we extract English corpora from SlimPajama~\cite{cerebras2023slimpajama}, excluding the book corpus, which was reported to have copyright infringement issues.
For the Japanese training data, we extract the Japanese portion of CommonCrawl corpus with the language identification and document deduplication applied using CCNet~\cite{wenzek-etal-2020-ccnet}.
For the vocabulary construction, we sample a small portion (50GB) from each language training data.

\section{Experiments on Vocabulary Size}
\label{sec:main_exp}

\subsection{Settings}
\label{sec:main_setting}

To investigate the relationship between vocabulary size and performance, we train Transformer-based language models on the training data described in Section \ref{sec:vocab_construction}.
Table \ref{tab:datasize} shows the number of tokens in the training data calculated from each vocabulary set.
As shown in this table, the number of tokens varies drastically based on the vocabulary size.
Therefore, we must take care not to give any unfair advantages to any setting.

For example, with a fixed number of training tokens, the 500k vocabulary model trains for around 1.5 epochs in English and 2 epochs in Japanese, while the 5k vocabulary model trains for only 1 epoch.
The larger vocabulary size has an advantage of seeing more data in this configuration.
In contrast, with a fixed number of training epochs, the 5k vocabulary model consumes much more computational resources than the larger vocabulary models.
Especially in Japanese, where the 5k vocabulary model contains about twice as much tokens as the 500k vocabulary model in 1 epoch.
Because the performance of LLMs is correlated with the computational costs during training~\cite{kaplan2020scaling}, this configuration might favor smaller vocabulary sizes.
Thus, we prepare two training configurations: 1T tokens and 1 epoch\footnote{In addition to the training data size, we have to discuss the number of parameters for a fair comparison because the model with the small vocabulary size contains less parameters for the embedding and output layers. However, as described in Appendix \ref{sec:comp_paramsize}, the model with the small vocabulary size does not improve the performance when we increase the number of parameters related to the vocabulary. Thus, we focus only on varying the training data size in our main experiments.}.

For hyper-parameters of the language model, we use the GPT-3 Large setting described in \citet{NEURIPS2020_1457c0d6}.
We set the number of layers $24$ and the hidden dimension size $1536$.
In this setting, the number of parameters for internal layers is 680M.
We use Megatron-LM~\cite{shoeybi2020megatronlm}\footnote{\href{https://github.com/NVIDIA/Megatron-LM}{https://github.com/NVIDIA/Megatron-LM}} as our codebase to train large language models.
To stabilize the training, we use the scaled embed technique~\cite{takase2024spike}.

We evaluate each model on the commonsense reasoning tasks.
For English, we use PIQA~\cite{Bisk_Zellers_Lebras_Gao_Choi_2020}, OpenBookQA (OBQA)~\cite{mihaylov-etal-2018-suit}, HellaSwag~\cite{zellers-etal-2019-hellaswag}, WinoGrande~\cite{10.1145/3474381} and ARC easy and challenge~\cite{clark2018think}.
For Japanese, we use JSQuAD and JCommonsenseQA (JCQA) from JGLUE~\cite{kurihara-etal-2022-jglue}, the Japanese portion of XWinograd~\cite{tikhonov-ryabinin-2021-heads}, and JAQKET\footnote{\href{https://sites.google.com/view/project-aio/competition1}{https://sites.google.com/view/project-aio/competition1}}.
Following the previous study~\cite{touvron2023llama}, we use the normalized likelihood in evaluation~\cite{NEURIPS2020_1457c0d6,eval-harness}.

\begin{table}[!t]
  \centering
  \footnotesize
  \begin{tabular}{ l | c c } \hline
  \multicolumn{1}{c|}{\#Vocab} & English & Japanese \\ \hline \hline
  5k & 830B & 950B \\
  10k & 750B & 750B \\
  50k & 670B & 590B \\
  100k & 650B & 550B \\
  500k & 640B & 490B \\ \hline
  \end{tabular}
  \caption{The number of tokens in training data tokenized by each vocabulary.}
  \label{tab:datasize}
\end{table}

\subsection{Results}

\begin{table*}[!t]
  \centering
  \footnotesize
  \begin{tabular}{ l | c c c c c c | c } \hline
  \multicolumn{1}{c|}{\#Vocab} & PIQA & OBQA & HellaSwag & WinoGrande & ARC-e & ARC-c & Avg.\\ \hline \hline
  \multicolumn{8}{c}{1T tokens} \\ \hline \hline
  5k & 69.9 & 33.2 & 51.0 & 55.2 & 49.6 & 27.7 & 47.8 ($\pm 0.0$)\\
  10k & 71.2 & 33.4 & 51.5 & 55.2 & 50.6 & 27.1 & 48.2 ($+ 0.4$) \\
  50k & \textbf{71.7} & 32.8 & 53.9 & 54.5 & 50.8 & 27.7 & 48.6 ($+0.8$)\\
  100k & 70.9 & 33.4 & 53.9 & 54.8 & 54.3 & 27.7 & 49.2 ($+1.4$)\\
  500k & 71.4 & \textbf{34.0} & \textbf{55.3} & \textbf{57.5} & \textbf{55.1} & \textbf{28.3} & \textbf{50.3} ($+2.5$)\\ \hline
  \multicolumn{8}{c}{1 Epoch} \\ \hline \hline  
  5k & 70.1 & 32.4 & 50.9 & \textbf{55.2} & 50.2 & \textbf{28.5} & 47.9 ($\pm 0.0$) \\
  10k & 71.1 & 33.6 & 50.6 & 55.7 & 49.0 & 27.1 & 47.9 ($\pm 0.0$) \\
  50k & 70.6 & 33.6 & 52.1 & 53.8 & 52.3 & 27.3 & 48.3 ($+ 0.4$)\\
  100k & \textbf{71.7} & 33.8 & 53.4 & 54.7 & 52.7 & 27.6 & 49.0 ($+ 1.1$)\\
  500k & 70.4 & \textbf{34.2} & \textbf{54.3} & 55.1 & \textbf{54.0} & 28.2 & \textbf{49.4} ($+ 1.5$)\\ \hline
  \end{tabular}
  \caption{The performance on English commonsense reasoning tasks in training 1T tokens and 1 epoch.}
  \label{tab:main_result_en}
\end{table*}

\begin{table*}[!t]
  \centering
  \footnotesize
  \begin{tabular}{ l | c c c c | c } \hline
  \multicolumn{1}{c|}{\#Vocab} & JSQuAD & JCQA & XWinograd & JAQKET & Avg. \\ \hline \hline
  \multicolumn{6}{c}{1T tokens} \\ \hline \hline
  5k & 58.1 & 68.1 & 58.9 & 12.5 & 49.4 ($\pm 0.0$) \\
  10k & 61.2 & 67.2 & 59.0 & 23.3 & 52.7 ($+ 3.3$) \\
  50k & 61.8 & 71.6 &  59.0 & 29.2 & 55.4 ($+ 6.0$) \\
  100k & 62.1 & \textbf{71.9} & \textbf{59.6} & 34.9 & 57.1 ($+ 7.7$) \\
  500k & \textbf{64.5} & 71.6 & 59.3 & \textbf{38.9} & \textbf{58.6} ($+ 9.2$) \\ \hline
  \multicolumn{6}{c}{1 Epoch} \\ \hline \hline
  5k & 57.7 & 68.1 & 58.8 & 14.4 & 49.8 ($\pm 0.0$) \\
  10k & 57.7 & 63.4 & \textbf{60.0} & 22.0 & 50.8 ($+ 1.0$) \\
  50k & 60.9 & 69.1 & 58.5 & 28.7 & 54.3 ($+ 4.5$) \\
  100k & 61.3 & \textbf{70.1} & 58.7 & 31.0 & 55.3 ($+ 5.5$) \\
  500k & \textbf{63.2} & 69.8 & 57.7 & \textbf{34.1} & \textbf{56.2} ($+ 6.4$) \\ \hline
  \end{tabular}
  \caption{The performance on Japanese commonsense reasoning tasks in training 1T tokens and 1 epoch.}
  \label{tab:main_result_ja}
\end{table*}



Tables \ref{tab:main_result_en} and \ref{tab:main_result_ja} present the performance of the models trained with 1T tokens and 1 epoch.
For each configuration, we show the average score of each task, and the improvement of the average score from the 5k vocabulary model.

As shown by the average scores, for both English and Japanese, larger vocabulary sizes lead to better performance.
The improvement is particularly notable in Japanese, largely due to the gains in JAQKET.
Unlike the other tasks where the model selects answers from provided candidates, JAQKET is a factoid QA task where the model generates answers without any candidates.
This suggests that a larger vocabulary size particularly benefits generation tasks.

In addition, the larger vocabulary size achieves better performance in either situation where we fix the number of training tokens or training epochs.
With a fixed number of epochs, the larger vocabulary size settings, e.g., 100k and 500k, use a much smaller number of training tokens (Table \ref{tab:datasize}).
This means that the larger vocabulary size also improves the training efficiency because we can obtain a better model with a smaller computational cost.
In fact, the GPU hours\footnote{We used A100 80GB for all experiments.} in the 100k setting are 0.7 times shorter than in the 5k when we fix the number of training epochs in Japanese\footnote{Since the larger vocabulary size slows the computation of the output distribution, we should use an efficient way such as the adaptive softmax~\cite{ICML-2017-GraveJCGJ} in practice.}.

\section{Experiments on Continual Training}
\label{sec:exp_continual_training}

\begin{table*}[!t]
  \vspace{-3.5mm}
  \centering
  \footnotesize
  \begin{tabular}{ l l | c c c c | l } \hline
  Setting & \multicolumn{1}{c|}{\#Vocab} & JSQuAD & JCQA & XWinograd & JAQKET & \multicolumn{1}{c}{Avg.} \\ \hline \hline
  From scratch & 100k & 71.8 & 76.0 & 63.6 & 54.2 & 66.4 \\ \hline
  Llama2 (w/o train) & 32k & 71.2 & 60.8 & 62.4 & 15.3 & 52.4 ($\pm 0.0$) \\
  Llama2 vocab & 32k & 80.7 & 79.4 & \textbf{72.6} & 47.7 & 70.1 ($+ 17.7$)\\
  Swap & 100k & 79.2 & \textbf{80.2} & 67.5 & 56.3 & 70.8 ($+18.4$)\\
  Swap\&Insert & 100k & \textbf{81.9} & \textbf{80.2} & 69.2 & \textbf{61.2} & \textbf{73.1} ($+ 20.7$)\\ \hline
  \citet{fujii2024continual} & 100k & 81.6 & 77.6 & 69.1 & 61.1 & 72.4 ($+ 20.0$) \\ \hline
  \end{tabular}
  \caption{The performance on Japanese commonsense reasoning tasks in the continual training from Llama2.}
  \label{tab:continual_training}
\end{table*}

\subsection{Increasing Vocabulary Size}
\label{sec:swap_method}
Section \ref{sec:main_exp} shows that the larger vocabulary size is useful in constructing LLMs from scratch.
In contrast, nowadays, we often start from a high-quality pre-trained model such as the Llama series~\cite{touvron2023llama} and continue training on the target language data~\cite{muller2022cedille,yong-etal-2023-bloom,yamada2024leia}.

Here, we check if we can readily increase the vocabulary size from the pre-trained model.
Similar techniques have been explored as vocabulary expansion~\cite{fujii2024continual,kim2024efficient} or sophisticated embedding initialization using cross-lingual word embeddings~\cite{minixhofer-etal-2022-wechsel}, but our focus here is to check if we could increase the vocabulary size in a rather simplistic way.
We consider a situation where we construct an entirely new vocabulary independently of the original vocabulary.

Let $V_{orig}$ and $V_{new}$ be the vocabulary set of the pre-trained model and a newly constructed vocabulary set respectively, and let $d$ be the dimension size of each layer.
To exploit knowledge learned in the pre-trained embedding matrix, we construct a new embedding matrix $E_{new} \in \mathbb{R}^{|V_{new}|\times d}$ from the original embedding matrix $E_{orig} \in \mathbb{R}^{|V_{orig}|\times d}$ with the way inspired by the randomized algorithm~\cite{doi:10.1137/090771806}:

\begin{align}
E_{new} = \frac{W E_{orig}}{\sqrt{|V_{orig}|}}, \label{eq:linear}
\end{align}
where $W \in \mathbb{R}^{|V_{new}| \times |V_{orig}|}$ is the random matrix whose elements are sampled from the standard normal distribution independently.
To maintain the standard deviation of $E_{orig}$ in $E_{new}$, we scale the matrix multiplication by $\frac{1}{\sqrt{|V_{orig}|}}$\footnote{We assume that $E_{orig}$ contains independent random variables with mean $0$ and variance $\mathrm{var}(E_{orig})$. Then, the variance of the matrix multiplication $W E_{orig}$ has mean $0$ and variance $\mathrm{var}(E_{orig}) \times |V_{orig}|$.}.

In the naive way, we swap $E_{new}$ with $E_{orig}$.
However, Equation \ref{eq:linear} randomizes embeddings even if $V_{new}$ contains the corresponding subwords which may possess useful knowledge transferable to the new model.
Therefore, we insert the pre-trained embedding in $E_{orig}$ into $E_{new}$ if the corresponding subword is included in both $V_{orig}$ and $V_{new}$\footnote{See Appendix \ref{sec:explanation_on_insert} for more details.}.
For the output layer, we construct a new weight matrix with the same manner.

\subsection{Results}
\label{sec:results_on_continual_training}
We train the Llama2 7B parameter model~\cite{touvron2023llama} with 100B tokens on our Japanese training data.
We use the Japanese vocabulary whose size is 100k.
Table \ref{tab:continual_training} shows results on Japanese commonsense reasoning tasks.
In this table, `Swap' uses new parameters related to the vocabulary without inserting the corresponding pre-trained parameters.
We train a language model from scratch to compare the effectiveness of the continual training.
Moreover, we compare the embedding initialization method by \citet{fujii2024continual} because their study is the same situation: continual training of Llama2 on Japanese data.

Table \ref{tab:continual_training} shows that `Swap' and `Swap\&Insert' outperform the model using the original Llama2 vocabulary even though these settings randomize parameters related to the vocabulary.
This result indicates that it is better to prepare an appropriate vocabulary even in the continual training situation.
Moreover, the insertion strategy achieves further improvement.
The `Swap\&Insert' outperforms the method of \citet{fujii2024continual}, which initializes an embedding of the new subword with the average of the pre-trained embeddings\footnote{For the existing subwords, their method uses the pre-trained embeddings. Thus, their method is regarded as using the `Insert' strategy.}, and thus, the `Swap\&Insert' is simple but effective.


\section{Related Work}
Before the paradigm of subword units and LLMs, researchers sometimes needed to handle the large vocabulary size such as more than 100k to decrease the number of unknown words.
For example, the vocabulary sizes of One Billion Word Benchmark and WikiText-103 are about 800k and 300k respectively~\cite{41880,DBLP:journals/corr/MerityXBS16}.
Some previous studies reported that character-level information was useful for neural language models with the large vocabulary size~\cite{45446,Takase:chara:ngram}.
In this paradigm, \citet{chen-etal-2019-large} explored the impact of the vocabulary size.

Since the use of subword units is proposed~\cite{sennrich-etal-2016-neural,kudo-2018-subword}, the vocabulary sizes 30k-60k are widely used as the magic numbers~\citep{libovicky-etal-2022-dont}.
As examples, the BERT and GPT papers use 30k and 40k for their vocabulary sizes respectively without any justification~\cite{NIPS2017_7181,devlin-etal-2019-bert,radford2018improving}.
\citet{kiyono-etal-2020-tohoku} investigated the relation between the performance and the vocabulary size but the maximum vocabulary size of their investigation is too small, i.e., 32k.

For large language models, the vocabulary sizes 30k-60k are also frequently used~\cite{noauthororeditor,touvron2023llama}.
In using the large vocabulary size, the authors claim to support multilinguality~\cite{le-scao-etal-2022-language,xue-etal-2021-mt5} or improve the efficiency~\cite{J1WhitePaper,llama3modelcard}.
In contrast, we investigate the relation between the vocabulary size and the performance of monolingual LLMs on each task.

\section{Conclusion}
In this paper, we empirically investigate the performance of monolingual LLMs when we vary the vocabulary size.
We conduct experiments on two languages: English and Japanese.
Experimental results show that the larger vocabulary size is, the better performance the language model achieves in both languages.
Moreover, we introduce a method to use the entirely new vocabulary in the continual training situation.
We show that using the appropriate vocabulary also improves the performance in the continual training.

\section*{Limitations}
In this study, we conducted experiments on two languages: English and Japanese.
We believe that our findings can be applied to other languages because we do not depend on linguistic features in the subword vocabulary construction.
However, we also agree that it is better to conduct exhaustive experiments on various languages to confirm the generality of our findings.

In this study, we used 500k as the maximum vocabulary size.
Because it is impractical to construct a much larger vocabulary than 500k, we could not investigate the improvement by the tremendously large vocabulary size such as one million and the upper bound of the performance.
The computational time of the vocabulary construction depends on the corpus size and the desired vocabulary size.
We roughly estimate that the vocabulary whose size is larger than one million requires at least over a month in its construction in our environment.

Furthermore, the parameter sizes of internal layers are 680M in training from scratch, and 7B in the continual training.
We consider that the discussions on subword vocabulary size are orthogonal to the parameter size of internal layers, but we would conduct additional experiments with more than 10B parameters if we had a large amount of computational resources.

\bibliographystyle{acl_natbib}


\clearpage

\onecolumn

\appendix

\section{Hyper-parameters}
\label{sec:hyperparam}

Table \ref{tab:hyper_params} shows hyper-parameters used in our main experiments described in Section \ref{sec:main_exp}.

\begin{table}[!t]
\centering
\footnotesize
\begin{tabular}{l | r} \hline
Hyper-parameter & Value \\ \hline \hline
Number of layers & 24 \\
Hidden dimension size & 1536 \\
Number of attention heads & 16 \\
Sequence length & 2048 \\
Batch size & 2048 \\
Learning rate & 3e-4 \\
Learning rate scheduler & Cosine \\
Warmup ratio & 0.01 \\
Adam $\beta_1$ & 0.9 \\
Adam $\beta_2$ & 0.95 \\
Weight decay & 0.01 \\
Gradient clipping & 1.0 \\ \hline
\end{tabular}
  \caption{Hyper-parameters used in experiments described in Section \ref{sec:main_exp}.}
  \label{tab:hyper_params}
\end{table}


\section{Formula of `Insert' in Section \ref{sec:swap_method}}
\label{sec:explanation_on_insert}
We formulate the procedure of `Insert' in Section \ref{sec:swap_method}.
Let $e_{i}^{orig}$ and $e_{i}^{new}$ be the $i$-th row vectors of $E_{orig}$ and $E_{new}$, and let $w_{i}^{orig}$ and $w_{i}^{new}$ be the corresponding subwords to $e_{i}^{orig}$ and $e_{i}^{new}$.
The `Insert' function, $\mathrm{Insert(\cdot)}$,  replaces $e_{i}^{new}$ with $e_{i}^{orig}$ when the corresponding subword is included in the original vocabulary $V_{orig}$ as follows:
\begin{align}
\mathrm{Insert}(e_{i}^{new}) = 
  \begin{cases}
  e_{j}^{orig} & \text{if $w_{i}^{new} \in V_{orig} \land w_{j}^{orig} = w_{i}^{new}$} \\
  e_{i}^{new} & \text{otherwise}
  \end{cases}
\end{align}
Therefore, the matrix contains both the randomized embeddings and the original pre-trained embeddings after the `Insert' procedure.
As shown in Section \ref{sec:results_on_continual_training}, this procedure leads to further improvement.

\section{Comparison on Vocabulary Expansion in Continual Training}
\label{sec:comp_vocab_expansion}

\begin{table*}[!t]
  \centering
  \footnotesize
  \begin{tabular}{ l l | c c c c | l } \hline
  Initialization & Vocab type & JSQuAD & JCQA & XWinograd & JAQKET & \multicolumn{1}{c}{Avg.} \\ \hline \hline
  \citet{fujii2024continual} & Expansion & 78.8 & 63.5 & 63.6 & 50.1 & 64.0 \\
  Swap\&Insert & Expansion & 80.7 & 60.6 & 67.4 & 55.9 & 66.2 \\
  \citet{fujii2024continual} & Appropriate & 81.6 & 77.6 & 69.1 & 61.1 & 72.4 \\
  Swap\&Insert & Appropriate & \textbf{81.9} & \textbf{80.2} & \textbf{69.2} & \textbf{61.2} & \textbf{73.1}\\ \hline
  \end{tabular}
  \caption{The performance on Japanese commonsense reasoning tasks in the continual training from Llama2 when we construct the 100k vocabulary with the vocabulary expansion approach and construct the appropriate 100k vocabulary to the Japanese training data.}
  \label{tab:comparison_vocab_expansion_appropriate}
\end{table*}

\begin{table}[!t]
  \centering
  \footnotesize
  \begin{tabular}{l | r} \hline
  \multicolumn{1}{c|}{Type} & Number \\ \hline \hline
  UTF-8 byte pieces & 256 \\
  Alphabet \& number (e.g., a, the, 1) & 5349 \\
  Symbol (e.g., +, =, \#\#) & 209 \\
  Others such as Japanese characters & 1083 \\ \hline \hline
  Total & 6897 \\ \hline
  \end{tabular}
  \caption{The type and number of shared subword units between the original Llama2 vocabulary and appropriate vocabulary, whose size is 100k, to the Japanese data in the continual training.}
  \label{tab:share_vocab_type}
\end{table}

In Section \ref{sec:exp_continual_training}, we conduct the continual training experiment on the scenario where we construct an appropriate vocabulary to the target language.
In this scenario, most subword units in the original vocabulary might be removed.
In contrast, the vocabulary expansion approach maintains the whole original vocabulary because it only adds new subword units to the original vocabulary~\cite{fujii2024continual}.
We investigate which approach is empirically better in this section.

We construct 100k vocabulary with the vocabulary expansion approach, and compare it with the appropriate vocabulary used in Section \ref{sec:exp_continual_training}.
We apply two strategies to initialize the embedding matrix: \citet{fujii2024continual} and our `Swap\&Insert'.
Table \ref{tab:comparison_vocab_expansion_appropriate} shows results of the continual training from Llama2.
This table indicates that using appropriate vocabulary outperforms the vocabulary expansion approach.
The appropriate vocabulary contains more subword units of the target language.
We consider that this property improves the performance.

Table \ref{tab:share_vocab_type} shows the shared subword units between the original Llama2 vocabulary and the appropriate vocabulary.
This table indicates that the number of shared subword units is only about 7000, which is about one-fifth of the original vocabulary.
Moreover, this table suggests that the original vocabulary contains few Japanese subword units because the number of the shared Japanese characters is about 1000.
Therefore, it is better to construct an entirely new vocabulary that is appropriate to the target language.

For the embedding initialization methods, Table \ref{tab:comparison_vocab_expansion_appropriate} shows that our `Swap\&Insert' achieves better averaged score than the method of \citet{fujii2024continual} in the same as the results in Section \ref{sec:exp_continual_training}.
Thus, our approach is also more suitable in the vocabulary expansion situation.

\section{Comparison on Parameter Size}
\label{sec:comp_paramsize}
The smaller vocabulary size lessens the parameter sizes related to the vocabulary in comparison with the larger vocabulary size.
Thus, the smaller vocabulary size might have the disadvantage in the number of parameters.
To confirm this point, we increase the parameters related to the vocabulary for the 5k setting.
Concretely, we expand the dimension of the embedding and output layers, and then modify the dimension size by the linear transformation such as the matrix factorization technique~\cite{Lan2020ALBERT:}\footnote{In contrast, we can reduce the number of parameters for the larger vocabulary size with the matrix factorization technique or more sophisticated way~\cite{NEURIPS2020_275d7fb2}, but we regard the 5k as the baseline in this experiment.}.
Let $|V|$ be the vocabulary size, $d_e$ be the dimension size of the embedding and output layers, and $d$ be the hidden dimension size.
We prepare the expanded embedding layer $E \in \mathbb{R}^{|V| \times d_e}$ and the trainable weight matrix $W \in \mathbb{R}^{d_e \times d}$.
We convert the dimension size of $E$ with the matrix multiplication $E W$.
For the output layer, we convert the dimension with the same manner.
We adjust $d_e = 30720$ for a fair comparison with the 100k setting in terms of the number of parameters.
For other hyper-parameters, we use the values shown in Table \ref{tab:hyper_params}.

Table \ref{tab:result_on_parameter_size} shows the performance on Japanese commonsense reasoning tasks.
This table indicates that the 5k with the expansion does not improve the average score although it increases the number of parameters.
This result suggests that the increase of the parameter size related to the vocabulary has no positive influence on the performance.
In contrast, the 100k achieves much better average score in the similar parameter size.
Therefore, the improvement by the increase of the vocabulary size is orthogonal to the increase of the parameter size.

\begin{table*}[!t]
  \centering
  \footnotesize
  \begin{tabular}{ l | c c | c c c c | c } \hline
  \multicolumn{1}{c|}{\#Vocab} & \multicolumn{1}{c}{Vocab \#Params.} & \multicolumn{1}{c|}{Total \#Params.} & JSQuAD & JCQA & XWinograd & JAQKET & Avg. \\ \hline \hline
  5k & \ \ \ \ 8M & 690M & 58.1 & 68.1 & 58.9 & 12.5 & 49.4 ($\pm 0.0$) \\
  5k w/ Expansion & 200M & 880M & 61.0 & 60.3 & 59.5 & 15.8 & 49.2 ($- 0.2$)\\
  100k & 150M & 840M & \textbf{62.1} & \textbf{71.9} & \textbf{59.6} & \textbf{34.9} & \textbf{57.1} ($+ 7.7$) \\ \hline
  \end{tabular}
  \caption{The performance on Japanese commonsense reasoning tasks when we use 1T tokens for training. For a fair comparison between 5k and 100k, we increase the parameter sizes of the embedding and output layers (Vocab \#Params. in this Table) for 5k with the matrix factorization technique~\cite{Lan2020ALBERT:}.}
  \label{tab:result_on_parameter_size}
\end{table*}

\section{Experiments on Each Training Data Size}
In addition to the 1T tokens in Section \ref{sec:main_exp}, we investigate the performance in other training data sizes: 10B, 50B, 100B, 200B, and 500B tokens.
Tables \ref{tab:each_data_size_en} and \ref{tab:each_data_size_ja} show the results of English and Japanese models when we use each training data size.
These tables show that larger vocabulary sizes lead to better performance for both English and Japanese in all training data sizes in the same as the results in Section \ref{sec:main_exp}.
These tables indicate that our findings are independent from the amount of training data.

The difference of the performance among vocabulary sizes is smaller in the 10B tokens than ones in other training data sizes.
Thus, the small training data size decreases the advantage of the large vocabulary sizes.
These results explain the relation between our findings and the previous study~\cite{ali-etal-2024-tokenizer}.
\citet{ali-etal-2024-tokenizer} concluded that the small vocabulary size such as 30k is sufficient for English monolingual LLMs.
We consider that they led the contrary conclusion to our findings because their training data, which is about 50B tokens, is much smaller than ours.

\begin{table*}[!t]
  \centering
  \footnotesize
  \begin{tabular}{ l | c c c c c c | c } \hline
  \multicolumn{1}{c|}{\#Vocab} & PIQA & OBQA & HellaSwag & WinoGrande & ARC-e & ARC-c & Avg.\\ \hline \hline
  \multicolumn{8}{c}{10B tokens} \\ \hline \hline
  5k & 58.4 & 25.4 & 29.3 & 51.9 & 34.3 & 22.3 & 36.9 ($\pm 0.0$) \\
  10k & 59.1 & 27.8 & 29.6 & \textbf{53.2} & 35.0 & 21.6 & 37.7 ($+ 0.8$) \\
  50k & 62.1 & 26.2 & 29.5 & 49.9 & 38.7 & 21.9 & 38.0 ($+ 1.1$) \\
  100k & \textbf{62.2} & \textbf{27.8} & 29.7 & 49.6 & \textbf{39.0} & 22.7 & 38.5 ($+ 1.6$) \\
  500k & 62.1 & 27.6 & \textbf{30.1} & 51.3 & 38.7 & \textbf{22.9} & \textbf{38.8} ($+ 1.9$) \\ \hline \hline
  \multicolumn{8}{c}{50B tokens} \\ \hline \hline
  5k & 66.7 & 28.0 & 39.0 & \textbf{52.3} & 41.9 & 23.8 & 41.9 ($\pm 0.0$) \\
  10k & 66.3 & 30.4 & 39.5 & 51.0 & 42.6 & 25.3 & 42.5 ($+ 0.6$) \\
  50k & 68.1 & 29.4 & 40.9 & 50.9 & 46.9 & 25.5 & 43.6 ($+ 1.7$) \\
  100k & 68.1 & 31.6 & 42.0 & 51.3 & 46.9 & 25.5 & 44.2 ($+ 2.3$) \\
  500k & \textbf{68.8} & \textbf{32.2} & \textbf{43.1} & 52.0 & \textbf{47.9} & \textbf{25.7} & \textbf{44.9} ($+ 3.0$) \\ \hline \hline
  \multicolumn{8}{c}{100B tokens} \\ \hline \hline
  5k & 67.2 & 30.8 & 42.7 & 52.2 & 44.1 & 26.7 & 44.0 ($\pm 0.0$) \\
  10k & 68.9 & \textbf{31.6} & 42.7 & 51.6 & 45.1 & 25.7 & 44.3 ($+ 0.3$) \\
  50k & 68.9 & 30.8 & 45.1 & 52.6 & 49.1 & 26.2 & 45.5 ($+ 1.5$)\\
  100k & 70.2 & \textbf{31.6} & 46.1 & 52.9 & 49.1 & 25.8 & 45.9 ($+ 1.9$) \\
  500k & \textbf{70.4} & \textbf{31.6} & \textbf{47.0} & \textbf{53.0} & \textbf{50.0} & \textbf{28.2} & \textbf{46.7} ($+ 2.7$) \\ \hline \hline
  \multicolumn{8}{c}{200B tokens} \\ \hline \hline
  5k & 68.8 & 32.8 & 45.3 & 53.4 & 46.0 & 25.3 & 45.2 ($\pm 0.0$) \\
  10k & 69.0 & 31.6 & 46.2 & 53.3 & 45.7 & 26.5 & 45.4 ($+ 0.2$)\\
  50k & 70.5 & 31.0 & 47.9 & 53.8 & 50.0 & 26.1 & 46.6 ($+ 1.4$)\\
  100k & 70.5 & \textbf{33.6} & 49.2 & \textbf{54.6} & 50.6 & 26.2 & 47.4 ($+ 2.2$)\\
  500k & \textbf{70.7} & 33.4 & \textbf{50.2} & 54.3 & \textbf{51.8} & \textbf{29.6} & \textbf{48.3} ($+ 3.1$) \\ \hline \hline
  \multicolumn{8}{c}{500B tokens} \\ \hline \hline
  5k & 69.7 & 32.6 & 49.6 & 52.9 & 47.7 & 26.4 & 46.5 ($\pm 0.0$) \\
  10k & 70.8 & \textbf{34.2} & 49.7 & 54.5 & 49.0 & 26.2 & 47.4 ($+ 0.9$) \\
  50k & 70.2 & 32.2 & 52.0 & 54.4 & 51.6 & 27.2 & 47.9 ($+ 1.4$)\\
  100k & 70.1 & 33.4 & 52.7 & 55.3 & 52.8 & 27.8 & 48.7 ($+ 2.2$) \\
  500k & \textbf{71.1} & 31.8 & \textbf{53.6} & \textbf{56.5} & \textbf{53.9} & \textbf{28.8} & \textbf{49.3} ($+ 2.8$) \\ \hline
  \end{tabular}
  \caption{The performance on English commonsense reasoning tasks when we use 100B, 200B, and 500B tokens for training.}
  \label{tab:each_data_size_en}
\end{table*}

\begin{table*}[!t]
  \centering
  \footnotesize
  \begin{tabular}{ l | c c c c | c } \hline
  \multicolumn{1}{c|}{\#Vocab} & JSQuAD & JCQA & XWinograd & JAQKET & Avg. \\ \hline \hline
  \multicolumn{6}{c}{10B tokens} \\ \hline \hline
  5k & 1.6 & 37.1 & 51.0 & 0.9 & 22.7 ($\pm 0.0$) \\
  10k & 1.4 & 44.2 & \textbf{53.6} & 0.5 & 24.9 ($+ 2.2$) \\
  50k & 2.7 & 47.9 & 51.0 & 1.6 & 25.8 ($+ 3.1$) \\
  100k & 5.3 & 48.9 & 51.7 & 3.3 & 27.3 ($+ 4.6$) \\
  500k & \textbf{10.1} & \textbf{50.8} & 52.5 & \textbf{4.1} & \textbf{29.4} ($+ 6.7$) \\ \hline \hline
  \multicolumn{6}{c}{50B tokens} \\ \hline \hline
  5k & 36.3 & 49.4 & 53.7 & 3.3 & 35.7 ($\pm 0.0$) \\
  10k & 42.6 & \textbf{59.1} & 56.0 & 7.7 & 41.4 ($+ 5.7$) \\
  50k & 42.8 & 56.8 & 55.7 & 12.2 & 41.9 ($+ 6.2$) \\
  100k & 40.9 & 56.8 & \textbf{56.9} & 17.5 & 43.0 ($+ 7.3$) \\
  500k & \textbf{48.9} & 57.2 & 54.8 & \textbf{17.9} & \textbf{44.7} ($+ 9.0$) \\ \hline \hline
  \multicolumn{6}{c}{100B tokens} \\ \hline \hline
  5k & 45.0 & 55.0 & 56.9 & 5.2 & 40.5 ($\pm 0.0$) \\
  10k & 49.8 & \textbf{60.9} & 56.7 & 12.3 & 44.9 ($+ 4.4$) \\
  50k & 51.4 & 56.0 & 56.8 & 18.4 & 45.7 ($+ 5.2$) \\
  100k & 49.1 & 58.7 & \textbf{58.9} & 20.7 & 46.9 ($+ 6.4$) \\
  500k & \textbf{56.3} & 60.1 & 55.7 & \textbf{26.3} & \textbf{49.6} ($+ 9.1$) \\ \hline \hline
  \multicolumn{6}{c}{200B tokens} \\ \hline \hline
  5k & 50.5 & 58.5 & 56.7 & 7.5 & 43.3 ($\pm 0.0$)\\
  10k & 53.8 & 61.1 & 57.7 & 16.9 & 47.4 ($+ 4.1$)\\
  50k & 55.8 & 54.0 & \textbf{58.7} & 21.4 & 47.5 ($+ 4.2$)\\
  100k & 54.7 & \textbf{64.2} & 58.0 & 27.2 & 51.0 ($+ 7.7$)\\
  500k & \textbf{60.6} & 61.4 & 56.4 & \textbf{32.1} & \textbf{52.6} ($+ 9.3$)\\ \hline \hline
  \multicolumn{6}{c}{500B tokens} \\ \hline \hline  
  5k & 56.4 & 62.2 & 58.4 & 10.4 & 46.9 ($\pm 0.0$) \\
  10k & 59.6 & 62.8 & 58.8 & 20.2 & 50.4 ($+ 3.5$)\\
  50k & 59.3 & \textbf{64.7} & 59.0 & 26.5 & 52.4 ($+ 5.5$)\\
  100k & 60.1 & \textbf{64.7} & \textbf{59.3} & 31.8 & 54.0 ($+ 7.1$)\\
  500k & \textbf{62.6} & 62.9 & 58.8 & \textbf{36.4} & \textbf{55.2} ($+ 8.3$)\\ \hline
  \end{tabular}
  \caption{The performance on Japanese commonsense reasoning tasks when we use 100B, 200B, and 500B tokens for training.}
  \label{tab:each_data_size_ja}
\end{table*}

\end{document}